\documentclass{article}
\usepackage{spconf}

\usepackage{psfrag,epsfig,graphics}
\usepackage{amsmath,amssymb,multirow}
\usepackage{subfigure}
\usepackage{mathtools}
\usepackage[below]{placeins}
\usepackage{cite}
\usepackage{algorithm}
\usepackage{algorithmic}
\ninept
\usepackage{colortbl}
\definecolor{Gray}{gray}{0.85}
\definecolor{LightCyan}{rgb}{0.88,1,1}

\newcommand{\cp}[1]{\ifmmode {\mathcal{#1}}\else ${\mathcal{#1}}$\fi}
\newcommand{\bzero}{\boldsymbol{0}}
\newcommand{\bone}{\boldsymbol{1}}

\newcommand{\bLambda}{\boldsymbol{\Lambda}}
\newcommand{\blambda}{\boldsymbol{\lambda}}
\newcommand{\ba}{\boldsymbol{a}}
\newcommand{\bA}{\boldsymbol{A}}
\newcommand{\bB}{\boldsymbol{B}}

\newcommand{\bC}{\boldsymbol{C}}
\newcommand{\bD}{\boldsymbol{D}}
\newcommand{\be}{\boldsymbol{e}}
\newcommand{\bE}{\boldsymbol{E}}
\newcommand{\bF}{\boldsymbol{F}}

\newcommand{\bG}{\boldsymbol{G}}

\newcommand{\bI}{\boldsymbol{I}}

\newcommand{\bL}{\boldsymbol{L}}
\newcommand{\bLa}{\mathcal{L}}

\newcommand{\br}{\boldsymbol{r}}
\newcommand{\bR}{\boldsymbol{R}}
\newcommand{\bs}{\boldsymbol{s}}
\newcommand{\bS}{\boldsymbol{S}}

\newcommand{\bv}{\boldsymbol{v}}
\newcommand{\bV}{\boldsymbol{V}}

\newcommand{\bW}{\boldsymbol{W}}
\newcommand{\bx}{\boldsymbol{x}}

\newcommand{\bX}{\boldsymbol{X}}

\newcommand{\bY}{\boldsymbol{Y}}
\newcommand{\bz}{\boldsymbol{z}}
\newcommand{\bZ}{\boldsymbol{Z}}
 
\newcommand{\tr}{\text{tr}}

\begin{document}

\title{A graph Laplacian regularization for hyperspectral data unmixing}

\name{Rita Ammanouil,  Andr\'e Ferrari, C\'edric Richard
\thanks{This work was partly supported by the Agence Nationale pour la Recherche, France, (Hypanema project, ANR-12-BS03-003), and the regional council of Provence-Alpes-C{\^o}te d'Azur.}
}
\address{Universit\'e de Nice Sophia-Antipolis, CNRS, Observatoire de la C\^ote d'Azur, France }

\maketitle

\begin{abstract}
This paper introduces a graph Laplacian regularization in the hyperspectral unmixing formulation. The proposed regularization relies upon the construction of a graph representation of the hyperspectral image. Each node in the graph represents a pixel's spectrum, and edges connect \emph{spectrally} and \emph{spatially} similar pixels. The proposed graph framework promotes smoothness in the estimated abundance maps and \emph{collaborative} estimation between homogeneous areas of the image. The resulting convex optimization problem is solved using the Alternating Direction Method of Multipliers (ADMM). A special attention is given to the computational complexity of the algorithm, and Graph-cut methods are proposed in order to reduce the computational burden. Finally, simulations conducted on synthetic data illustrate the effectiveness of the graph Laplacian regularization with respect to other classical regularizations for hyperspectral unmixing.
\end{abstract}



\section{Introduction}

Hyperspectral sensors provide both a spatial and a spectral representation of a scene. They acquire images throughout the visible and Infrared  portions of the spectrum, with a spectral resolution as narrow as $1$ nm. Depending on the working distance of the hyperspectral camera, the spatial resolution can be of a few micrometers~(laboratory measurements) up to a few meters (airborne remote sensing). As a result every pixel in the hyperspectral image is a vector of reflectance values also known as the pixel's spectrum. Unmixing \cite{Keshava02} is one of the most prominent tools to analyze hyperspectral data. It consists of identifying the pure components in the captured scene, the so-called endmembers, and then estimating their spatial distributions, also known as their abundance maps. Most unmixing methods in the literature focus on the Linear Mixing model \cite{Heinz01}, where each pixel is modeled by a convex combination of the endmembers weighted by their abundances. 


The purpose of this paper is to introduce the graph Laplacian regularization in the hyperspectral unmixing formulation. This is motivated by the intuition that pixels with {similar} spectral structure and similar spatial contextual information will have broadly similar abundances. Representing these pairwise {similarities} by edges gives rise to a graph, where each node represents a pixel. The resulting graph structure provides additional {relational} information which can improve the abundance estimation accuracy and complement existing \emph{pixel-by-pixel} unmixing techniques. As we shall see further ahead in Section \ref{sec:LapRegUn}, the graph Laplacian regularization provides an elegant and flexible way to incorporate this information in the unmixing  problem by means of a closed form expression for penalizing the difference between the estimates of similar pixels via the $\ell_2$-norm. This regularization has been widely used in many fields especially in semi-supervised learning also known as transductive learning. The potential of this regularization has been demonstrated for many applications including digit recognition and text classification \cite{Zhou2004}, web-page categorization \cite{Zhang2006}, hyperspectral data classification \cite{Camps2007}, manifold learning \cite{Belkin2006}, and image denoising \cite{Kovac2011} to cite a few.


The proposed strategy is closely related to the work in \cite{Iordache2012} where the authors use a Total Variation (TV) regularization on top of sparse $\ell_1$-norm regularized unmixing.  Similarly to \cite{Iordache2012}, this communication advocates the use of the graph Laplacian regularization on top of $\ell_{21}$-norm regularized unmixing. TV is restricted to the assumption of local spatial similarity, and assumes that a pixel is only similar to its four neighbors. However, the graph Laplacian regularization is more flexible in the sense that it allows to connect a pixel with as many other pixels in the image as long as they are similar. \cite{Chen2014} extends the TV spatial regularization to nonlinear unmixing models. Several methods in the literature incorporate other spatial or spectral-spatial information in the unmixing problem such as \cite{Jia2007,Castrodad2011,Eches2011,Zare2011}. For a detailed review of spectral unmixing methods and endmember extraction techniques with spatial information, the reader is referred to \cite{Shi2014}. Very recently, the authors of \cite{Tong2014} used the graph Laplacian regularization on top of sparse $\ell_{1/2}$ nonnegative matrix factorization (NMF) for blind unmixing. The algorithm uses alternate minimization in order to simultaneously estimate the endmembers and the abundances. In this work, we use the ADMM algorithm \cite{Boyd10} which allows to take into account the abundances sum-to-one and positivity constraints, and a Group lasso regularizer frequently incorporated in unmixing to allow the use of large libraries of endmembers
\cite{Iordache2013}. Moreover, we exploit the graph structure and use an algorithm \cite{Ng2002} similar to normalized graph-cuts \cite{Shi2000} in order to partition the graph into several sub-graphs. Unmixing is then performed on each subgraph separately which allows to reduce the computational complexity of the algorithm.

The paper is organized as follows. Section \ref{sec:GraphMap} introduces the hyperspectral and graph frameworks, Section \ref{sec:LapRegUn} incorporates the graph Laplacian regularization on top of sparse unmixing, Section \ref{sec:exp} is devoted to testing the proposed approach using synthetic data. Finally, Section \ref{sec:conclusion} concludes the paper.








 
\section{Hyperspectral image to graph mapping} 
\label{sec:GraphMap}

Let us first introduce the linear mixing model and some notations specific to the hyperspectral unmixing framework. In matrix form, the linear mixing model is given by
\begin{equation} 
	\label{LinearModel}
	\bS = \bR{\bA} + \bE
\end{equation}
with  $\bS = (\bs_{1},\ldots, \bs_{N}) $, $\bR = (\br_{1},\ldots, \br_{M})$, ${\bA} = ({\ba}_{1},\ldots, {\ba}_{M})^{\top}$. Here, $\bs_{j}$ is the  $L$-dimensional  spectrum of the $j$-th pixel, $L$ is the number of frequency bands, $\br_{i}$ is ${L}$-dimensional  spectrum of the $i$-th endmember, $M$ denotes the number of endmembers, ${\ba}_i$ is the ${N}$-dimensional abundance map of the $i$-th endmember, $N$ is the number of pixels in the image, and $\bE$ is an additive Gaussian noise. In addition, let $\bs_{\lambda_i}$ be the $i$-th row of $\bS$ that denotes the collection of the $N$ spectrum values at the $i$-th spectral band of $\bS$. All vectors are column vectors. Model \eqref{LinearModel} means that the $(i,j)$-th entry ${A}_{ij}$ of $\bA$ is the abundance of the endmember $\br_{i}$ in pixel $\bs_{j}$. Two constraints on the abundances are usually considered, the non-negativity and sum-to-one constraints: $A_{ij} \geq 0 $ for all $(i,j)$, and $ \sum_{i=1}^M A_{ij} = 1$ for all $j$.

The first step in the proposed graph-based unmixing approach consists in mapping the hyperspectral image to a graph $\bG$ where each node represents a pixel's spectra. Let $\bW$ be the $N \times N$ affinity matrix of the graph, the entries $W_{ij}$ of $\bW$ satisfy the following conditions. If pixels $i$ and $j$ are similar then $W_{i,j}$ is set to some positive value proportional to their degree of similarity. If pixels $i$ and $j$ are dissimilar then $W_{ij}$ tends to zero. There are different heuristics for choosing $W_{ij}$. For example, this can be done by using a Gaussian kernel
\begin{equation}
	W_{ij}=\exp\left(-\frac{\|\bs_i -\bs_j\|^2}{2 \sigma^2}\right)
\end{equation}
where $\sigma$ is the kernel's bandwidth \cite{Gillis2012,Zhang2014}. In addition to the pixel's spectrum, each pixel can be defined by a vector of spatial features, for instance, the average of its surrounding area, its coordinates in the image. This spatial information leads to a second spatial affinity matrix which can be easily combined with the spectral one \cite{Camps2007}. Finally, k-nearest neighbors and thresholding are commonly used in order to set to zero small weights in $\bW$ \cite{Argyriou2005}. The authors of \cite{Zhang2014,Gillis2012,Camps2007} propose different strategies for defining an affinity matrix that takes into account both the spatial and the spectral information of a pixel.

\section{Laplacian regularized unmixing}
\label{sec:LapRegUn}
As previously mentioned, we consider the following interpretation of the graph. If two nodes are connected, then they are likely to have similar abundances. We shall now incorporate this information in the unmixing problem using the graph Laplacian regularization. This leads to the following convex optimization problem: 
\begin{equation} \label{cvx1}
	\begin{array}{ll} 
	\min_{\bA}   	& \frac{1}{2}\|\bS-\bR\bA\|_\text{F}^2  + \lambda \tr(\bA \bLa \bA^\top) + \mu \sum_{k=1}^N  \| \ba_{k}\|_2\\
	\text{subject to} & A_{ij} \geq 0 \quad \forall\, i,j \\ 
			&  \sum_{i=1}^N A_{ij} = 1 \quad \forall\, j.
	\end{array}
\end{equation} 
where $\bLa$ is the graph Laplacian matrix given by $\bLa = \bD - \bW$, $\bD$ is a diagonal matrix with $\bD_{ii} = \sum_{j=1}^{N}W_{ij}$, $\mu \geq 0$ and $\lambda \geq 0$ are two regularization parameters. The first term in \eqref{cvx1} is a data fidelity term based on the $\ell_2$-norm. The second term is the graph Laplacian regularization. To see the relevance of this regularization in \eqref{cvx1}, we rewrite it as follows \cite{Mohar1991}:
\begin{equation}
\label{Lapl}
	\tr(\bA \bLa \bA^\top) = \sum_{i=1}^M \sum_{j=1}^N \sum_{k \sim j} W_{jk} (A_{ij}-A_{ik})^2
\end{equation}
where $k \sim j$ indicates that pixels $j$ and $k$ are similar ($W_{jk} \neq 0$). For every abundance map (row in $\bA$), this term penalizes the square of the difference between the abundances of similar pixels proportionally to their degree of similarity. This quantity can also be seen as a measure of the discrepancies between the abundance estimates weighted by their degree of similarity $W_{jk}$. The regularization parameter $\lambda$ controls the extent at which similar pixels estimate similar abundances. The third term is the $\ell_{21}$-norm regularization also known as the Group lasso. We consider that $\bR$ is a large dictionary of endmembers, and only few of these endmembers are present in the image. For this reason, we use the the Group Lasso regularization to induce group sparsity \cite{Yuan06} in the estimated abundance matrix by possibly driving several rows $\ba_k$ of $\bA$ to zero, as proposed in \cite{Iordache2013}.

It is important to note that the first and second term of the cost function \eqref{cvx1} can be grouped in a single quadratic form. However the resulting Quadratic Problem has $N \times M$ non-separable variables. The transpose being on the second $\bA$ instead of the first one in \eqref{Lapl} makes the problem non-separable with respect to the columns of $\bA$. The solution of problem \eqref{cvx1} can be obtained in a simple and flexible manner using the Alternating Direction Method of Multipliers \cite{Boyd10}. 

\subsection{ADMM algorithm}
We consider the canonical form and the following variable splitting:
\begin{equation} \label{cvx2}
	\begin{array}{ll} 
	\min_{\bX,\bY,\bZ} & \frac{1}{2}\|\bS-\bR\bX\|_\text{F}^2 + \lambda \tr(\bY \bLa \bY^\top) + \mu \sum_{k=1}^N  \|\bz_{k}\|_2 \\
	           & \quad +  \cp{I}(\bZ)  \\
	\text{subject to} & \bB\bX + \bC\bZ = \bF \\
	                          & \bX = \bY
	\end{array}
\end{equation} 
with
\begin{equation}
	\bB = \left(\begin{array}{c} \bI\, \\ \bone^ \top \end{array}\right),\;
	\bC = \left(\begin{array}{c} -\bI \\ \,\bzero^\top\end{array}\right),\;
	\bF = \left(\begin{array}{c} \bzero\, \\  \bone^\top  \end{array}\right),  \nonumber
\end{equation}
where $\cp{I}$ is the indicator of the positive orthant guarantying the positivity constraint, that is, $\cp{I}(\bZ)= 0$ if $\bZ\succeq\bzero$ and $+\infty$ otherwise. The constraints impose the consensus $\bX=\bY$, $\bX=\bZ$, and the sum-to-one. In matrix form, the augmented Lagrangian for problem~\eqref{cvx2} is given by
\begin{equation}
\begin{split}
	{\bL}_\rho( & \bX,\bY,  \bZ,\bV,\bLambda) = \frac{1}{2}\|\bS-\bR\bX\|_F^2+ \mu \sum_{k=1}^N  \|\bz_{k}\|_2 +\cp{I}(\bZ)  \\
	&+ \lambda \tr(\bY \bLa \bY^\top) + \tr(\bV^\top (\bX - \bY))+ \frac{\rho}{2}\, \| \bX  - \bY\|_\text{F}^2 \\
	& + \frac{\rho}{2}\, \| \bB\bX + \bC\bZ - \bF\|_\text{F}^2  + \tr(\bLambda^\top (\bB\bX + \bC\bZ - \bF)) 
\end{split}
\end{equation}
where $\bLambda$ and $\bV$ are the matrices of the Lagrange multipliers, and $\rho$ is the penalty parameter. The flexibility of the ADMM lies in the fact that it splits the initial optimization problem into three subproblems. At iteration $k+1$, the ADMM algorithm is outlined by four sequential steps.

\medskip 

\noindent \textbf{$\bX$ minimization step}: The augmented Lagrangian is quadratic with respect to $\bX$. The minimizer has an analytical expression that is obtained by setting the gradient of the augmented Lagrangian with respect to $\bX$ to zero:
\begin{equation}
	\begin{split}
	\bX^{k+1}   &  = ({\bR}^\top \bR+\rho \bB^\top\bB + \rho \bI_N)^{-1} (\bR^\top\bS \\
	   & - \bB^\top [\bLambda^k+\rho\,(\bC\bZ^k-\bF)] -\bV^k + \rho \bY^k).
	\end{split}
\end{equation}

\medskip

\noindent\textbf{$\bY$ minimization step}: Similarly to the first step, $\bY^{k+1}$ is obtained by setting the gradient of the augmented Lagrangian with respect to $\bY$ to zero, which yields:
\begin{equation}
	\begin{split}
	\bY^{k+1}   &  = (\bV^k + \rho\bX^{k+1})(2\lambda\bLa + \rho \bI)^{-1}.
	\end{split}
\end{equation}
Assume that we did not use $\bY$, and assigned the same ADMM variable $\bX$ for both the fidelity term and the graph Laplacian regularization. In this case, the $\bX$ minimization reduces to solving a Sylvester equation \cite{Bartels72}. The exact solution of this problem can not be computed efficiently due to the high dimensionality of the problem. 
In fact it requires the inversion of a $NM\times NM$ matrix where $N$ and $M$ can be both very large. Iterative methods have been proposed to perform this task \cite{Ding2005}. These iterative methods are similar to the first two steps of our ADMM solution in the sense that the initial variable is split into two variables and alternating updates of these variables are performed. 

\medskip
\noindent \textbf{$\bZ$ minimization step}: 
After {discarding} the terms that are independent of $\bZ$, the minimization of the augmented Lagrangian with respect to $\bZ$ reduces to solving the following problem:
\begin{equation}
	\label{ADMMz}
	\begin{array}{ll} 
		\min_{\bZ}   & \mu \sum_{k=1}^N  \|\bz_{k}\|_2 +\tr(\bLambda^\top \bC\bZ) \\
		 & +\frac{\rho}{2} \| \bB\bX + \bC\bZ - \bF\|_\text{F}^2\\
		\text{subject to} & \bZ\succeq\bzero.
	\end{array}
\end{equation}
This minimization step can be split into $N$ problems given the structure of matrices $\bB$ and $\bC$, one for each row of $\bZ$, that is,
\begin{equation}
	\label{zmin}
	\begin{array}{ll} 
		\min_{\bz}   & \frac{1}{2} \|\bz-\bv\|_2^2 + \alpha  \|\bz\|_2 + \cp{I}(\bz)
	\end{array}
\end{equation} 
where $\bv = \bx +\rho ^{-1} \blambda$, $\alpha = \rho ^{-1}\mu$. {Vectors} $\blambda$, $\bx$ and $\bz$ correspond to a given row {of} $\bLambda$, $\bX$ and $\bZ$, respectively. The minimization problem~\eqref{zmin} admits a unique solution given by the proximity operator of function $f(\bz)=\alpha  \|\bz\|_2  + \cp{I}(\bz)$:
\begin{equation}
	\label{Pmisto}
	\left\{
	\begin{array}{ll}
		\bz^\ast = \bzero  & \text{if } \|(\bv)_+\|_2 < \alpha  \\
		\bz^\ast  = \left(1 - \frac{\alpha}{\|(\bv)_+\|_2}\right) (\bv)_+  & \text{otherwise},
	\end{array}\right.
\end{equation}
where $(\cdot)_+ = \max(\bzero,\cdot)$. Operator \eqref{Pmisto} was recently used in \cite{Thiebaut13,Ammanouil2014}. {The derivation of} this operator can be found in \cite{Ammanouil2014}. 

\medskip

\noindent \textbf{Update of the Lagrange multipliers $\bLambda$ and $\bV$}: The last step consists of updating the Lagrange multipliers $\bLambda$ and $\bV$ using the following expressions

\begin{equation}
\begin{array}{c}
	\bLambda^{k+1} = \bLambda^{k} +\rho(\bB\bX^{k+1}+\bC\bZ^{k+1}-\bF),	\\
	\bV^{k+1} = \bV^{k} +\rho(\bX^{k+1}-\bY^{k+1}).	
\end{array}
\end{equation}
As suggested in \cite{Boyd10}, a reasonable stopping criteria for this iterative algorithm is that the primal and dual residuals must be smaller than some tolerance thresholds.

\subsection{A note on complexity}
Finally, we pay particular attention to the computational complexity of the resulting ADMM algorithm. The most expensive step is the $\bX$ minimization, since it requires the inversion of an $N \times N$ matrix, $N$ being very large in real images. We propose to exploit the graph representation of the pixels and apply the algorithm of \cite{Ng2002} in order to partition the nodes of the graph into $k$ clusters or subgraphs. Then the ADMM algorithm can be applied on each subgraph where the number of pixels is now smaller than $N$. The affinity matrix of each subgraph is a subset of the original graph affinity, therefore it does not need to be recomputed. 

The purpose of this step is to reduce the computational complexity while preserving the global knowledge captured by the graph structure. For this reason the segmentation must be ``conservative''. 

\section{Experiments}
\label{sec:exp}
\begin{figure*}
\centering

\subfigure[$\be_{2}$: True]{ \includegraphics[trim = 10cm 5cm 13cm 3cm, clip = true,width=0.24\textwidth, height = 0.13\textheight]{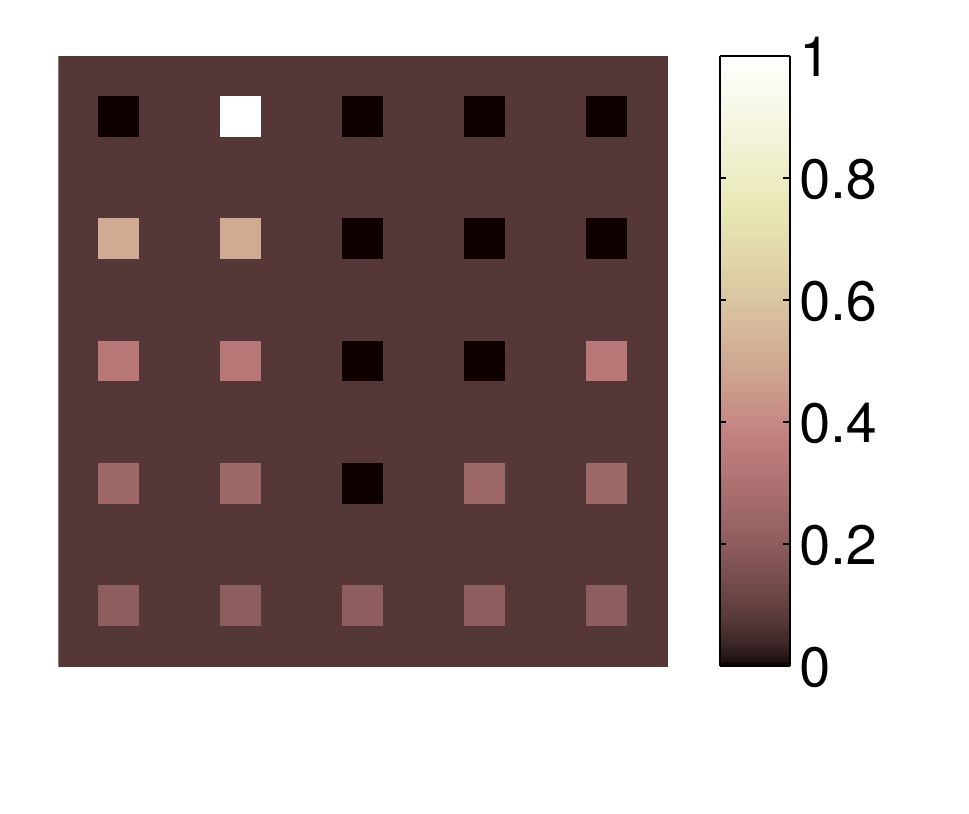}} 
\subfigure[$\be_{2}$: FCLS]{ \includegraphics[trim = 10cm 5cm 13cm 3cm, clip = true,width=0.24\textwidth, height = 0.13\textheight]{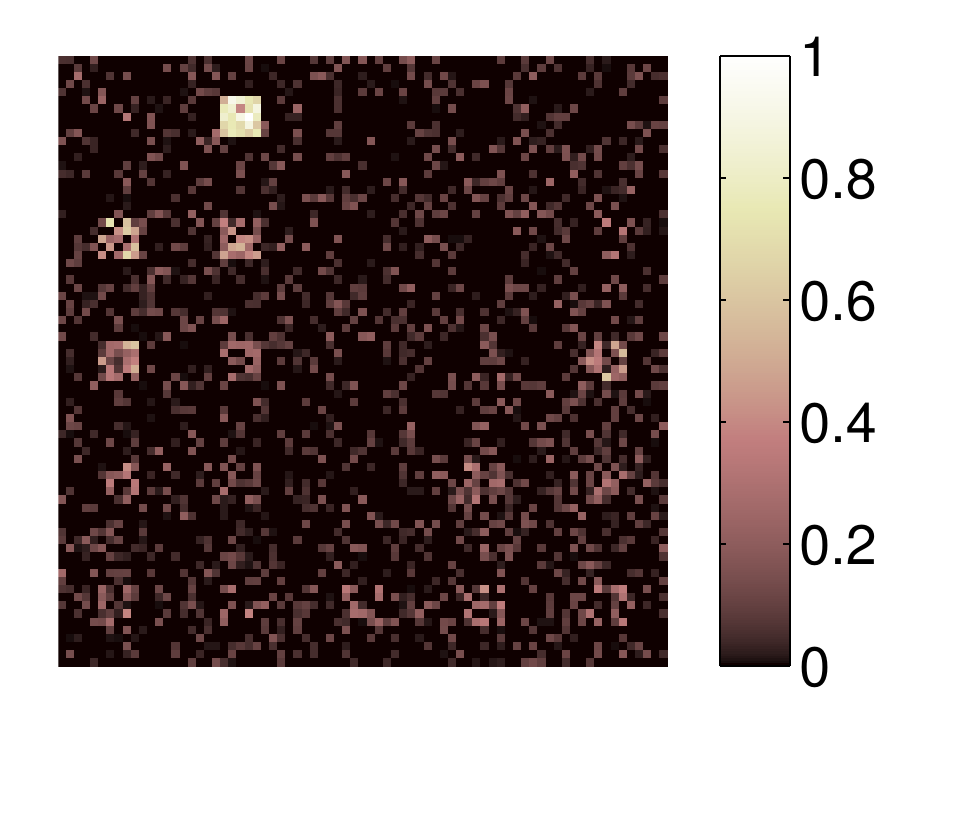}}
\subfigure[$\be_{2}$: SUnSAL-TV]{ \includegraphics[trim = 10cm 5cm 13cm 3cm, clip = true,width=0.24\textwidth, height = 0.13\textheight]{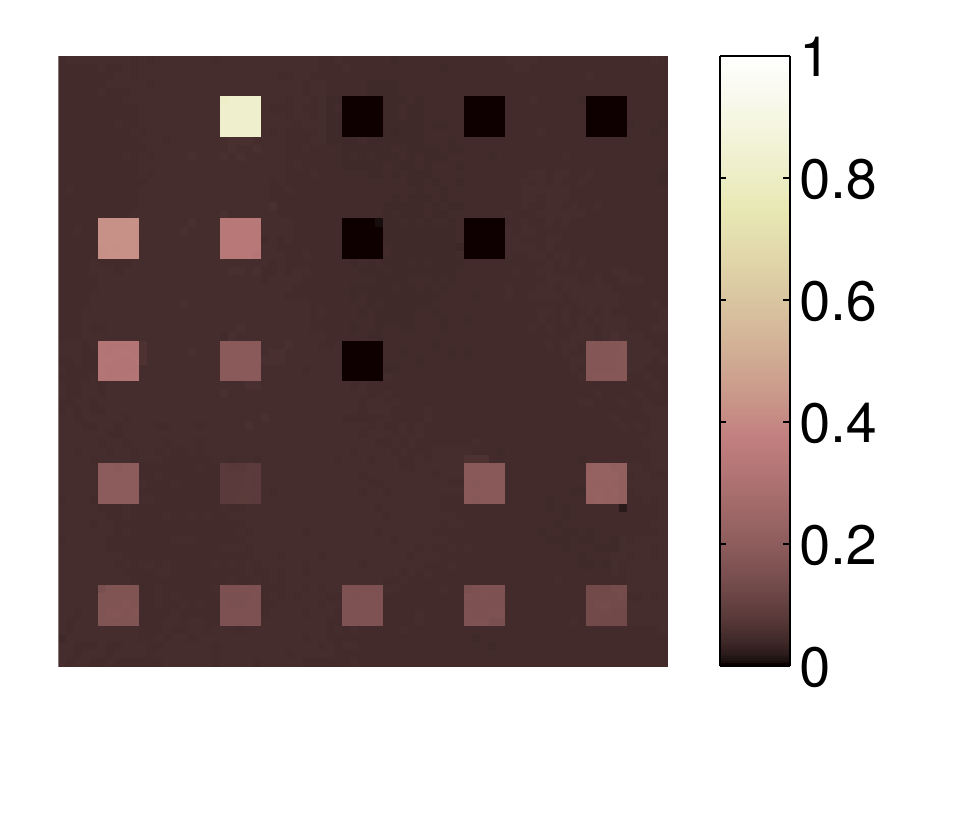}}
\subfigure[$\be_{2}$: GLUP-Lap]{ \includegraphics[trim = 10cm 5cm 13cm 3cm, clip = true,width=0.24\textwidth, height = 0.13\textheight]{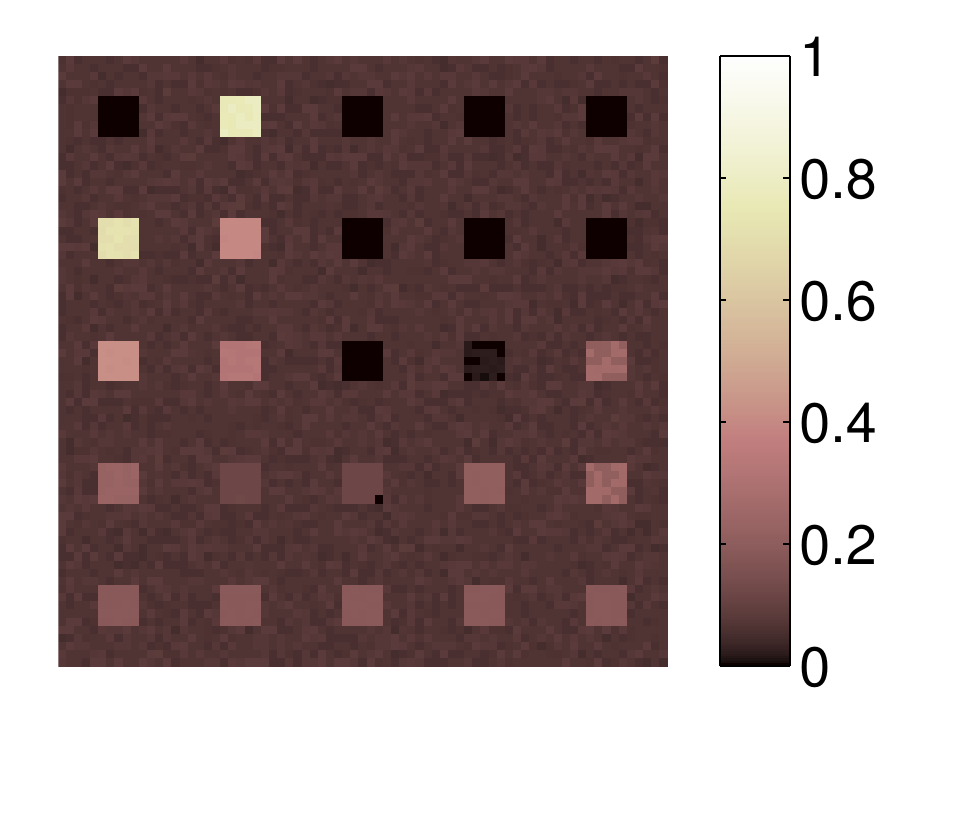}} \\

\subfigure[$\be_7$: True ]{ \includegraphics[trim = 10cm 5cm 13cm 3cm, clip = true,width=0.24\textwidth, height = 0.13\textheight]{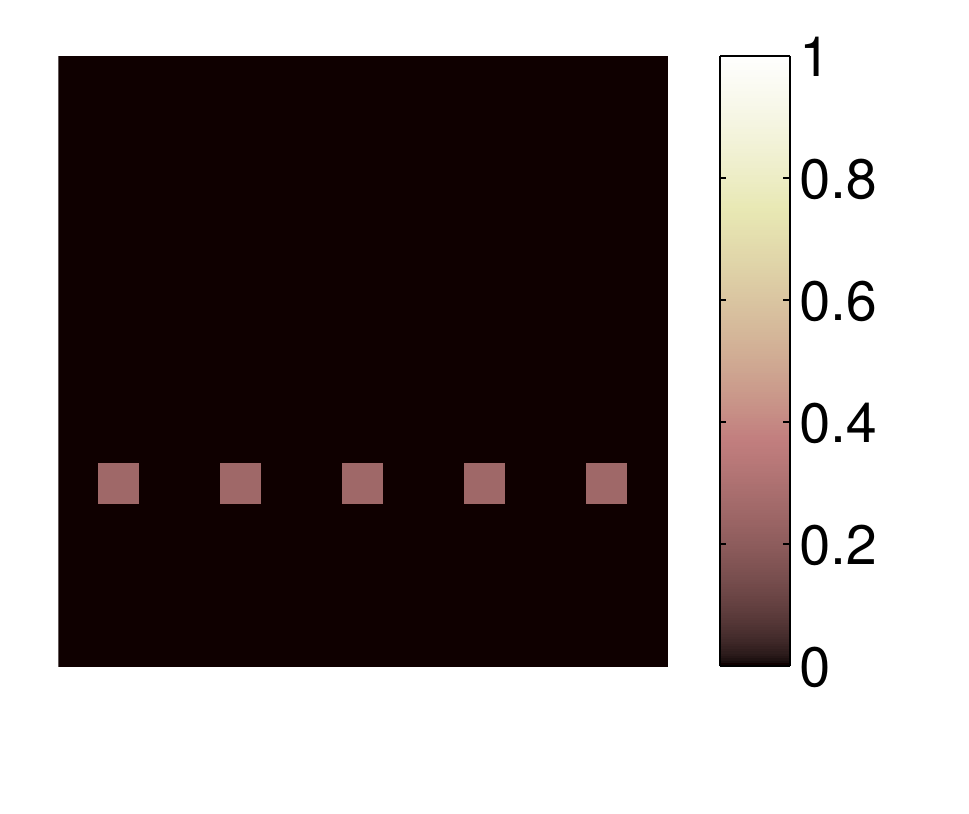}}
\subfigure[$\be_7$: FCLS ]{ \includegraphics[trim = 10cm 5cm 13cm 3cm, clip = true,width=0.24\textwidth, height = 0.13\textheight]{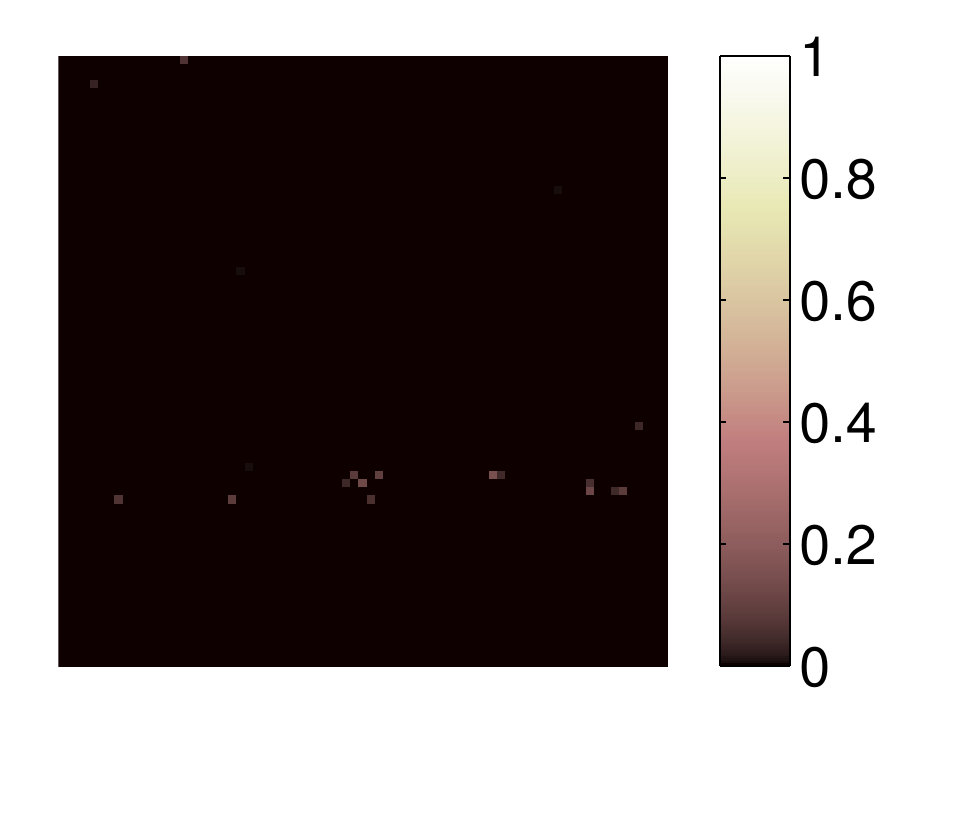}}
\subfigure[$\be_7$: SUnSAL-TV]{ \includegraphics[trim = 10cm 5cm 13cm 3cm, clip = true,width=0.24\textwidth, height = 0.13\textheight]{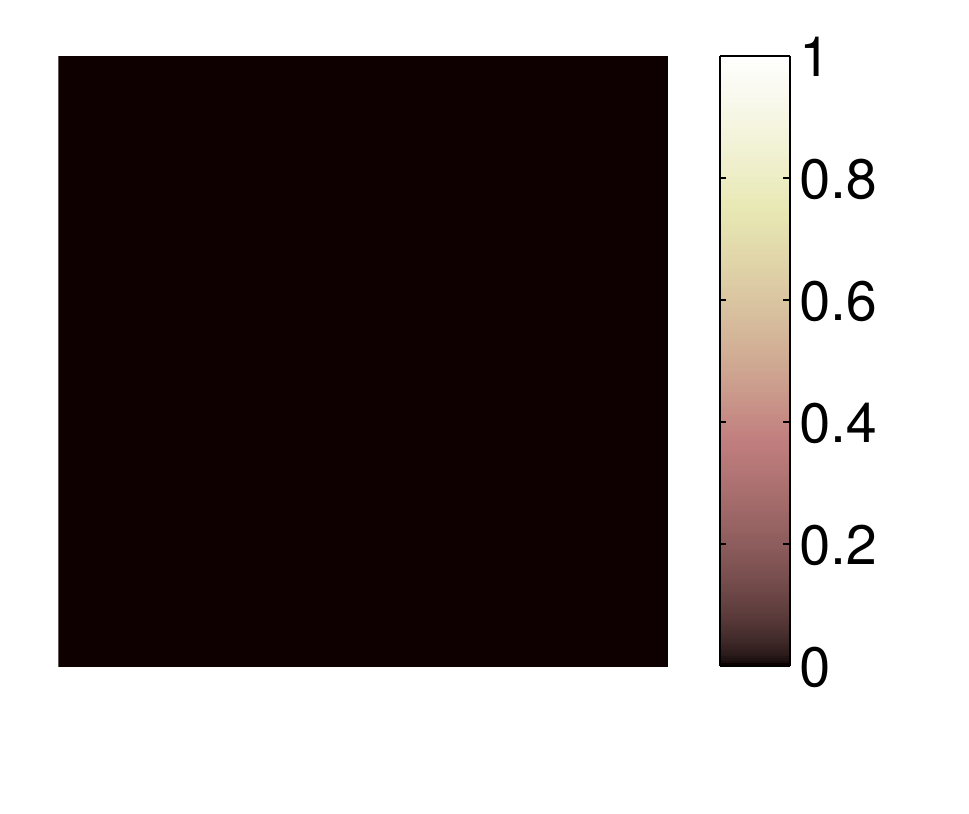}}
\subfigure[$\be_7$: GLUP-Lap]{ \includegraphics[trim = 10cm 5cm 13cm 3cm, clip = true,width=0.24\textwidth, height = 0.13\textheight]{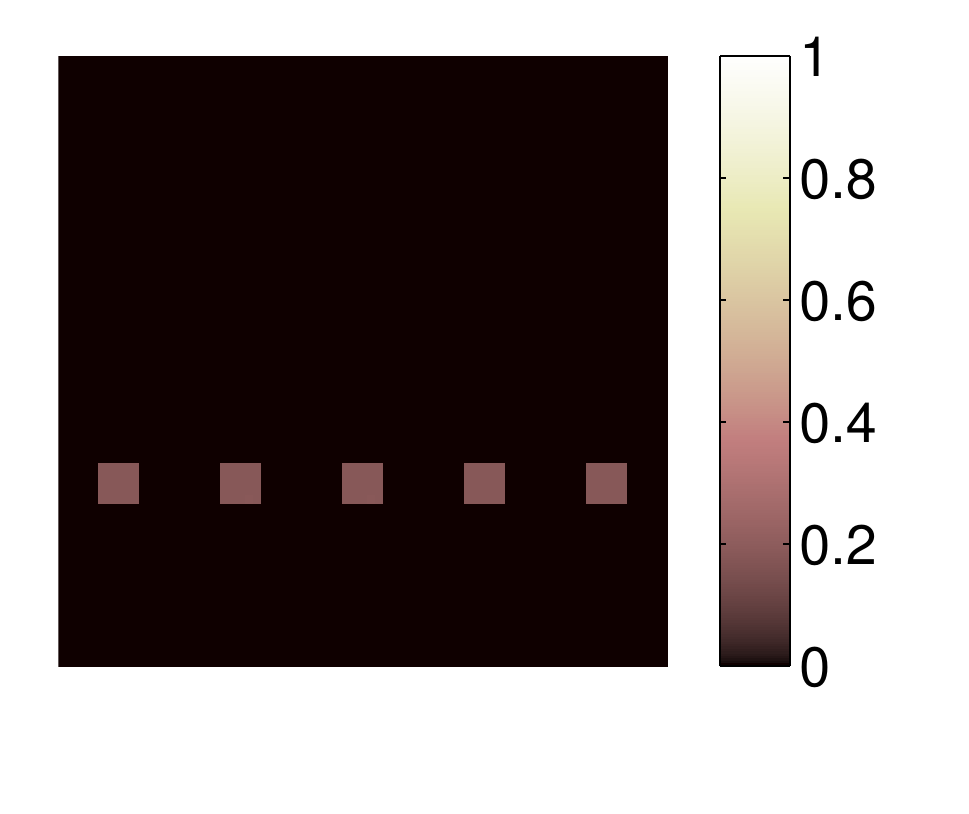}} \\

\caption{First row: Abundance maps for endmember $2$ in Data$1$ obtained with $\text{SNR}=30$ dB. Second row: Abundance maps for endmember $7$ in Data$2$ obtained with $\text{SNR}=30$ dB. From left to right: The true abundance map, FCLS, SUnSAL-TV, GLUP-Lap. The parameters are the reported in Table 1.}
\label{AbdMaps}
\end{figure*}

The performance of the proposed approach was evaluated using two simulated data sets, namely, Data$1$ and Data$2$ designed with different levels of homogeneity. Data$1$ is the same data set used in the experiments of \cite{Iordache2012,Chen2014}. The image consists of $75\times75$ pixels generated using $5$ endmembers $[\be_1, \be_2, \cdots \be_{5}]$ with $224$ bands extracted from the USGS library. The background of the image is a mixture of the $5$ endmembers with the following abundances $[0.1149\, 0.0741\, 0.2003\, 0.2055\, 0.4051]^{\top}$. There are $25$ squares in the image disposed in a $5\times5$ grid fashion (see Figure \ref{AbdMaps}). Each square is an homogeneous surface where its pixels have the same abundances. The first $20$ squares are different from each other, each one contains a different mixture of the endmembers, whereas the last $5$ horizontally aligned squares are identical. Most of the abundances in Data$1$ verify the assumption of local consistency. Data$2$ is generated similarly to Data$1$, except that it is created using $15$ distinct endmembers, and the squares in each row are identical. In addition to local consistency, there exists distant homogeneous surfaces in Data$2$ that are identical. As a result a pixel has local similar neighbors and distant ones too. 
 

\begin{figure}
\centering
{ \includegraphics[trim = 0.5cm 0cm 4cm 0cm, clip = true,width=0.35\textwidth, height = 0.22\textheight]{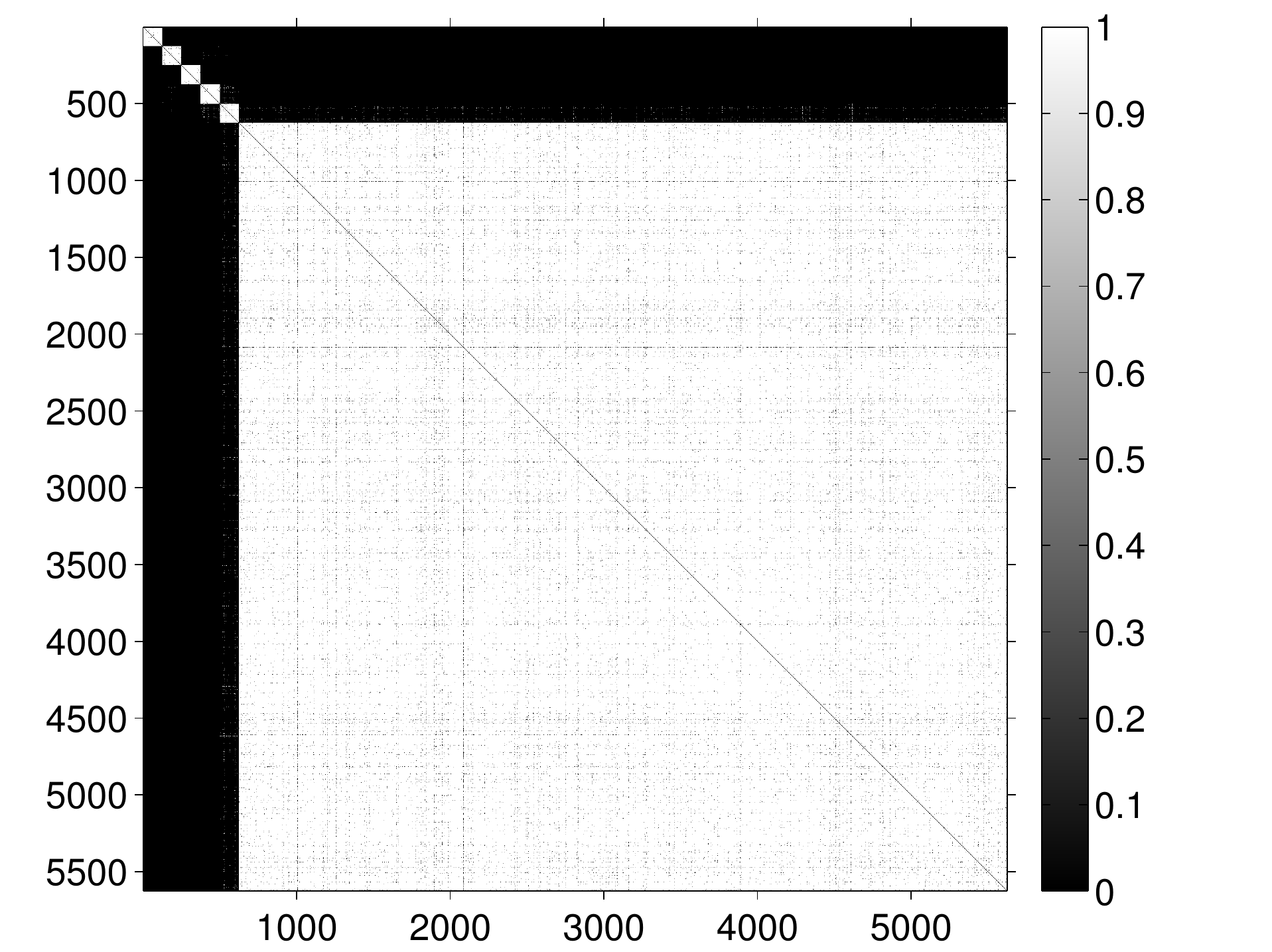}}
\caption{Affinity matrix of Data$2$ obtained with $\text{SNR}=20 ~ \text{dB}$ and $d_{\min}^2=1.8$.}
\label{TrueImage}
\end{figure}

The first step in the proposed approach consists of defining the affinity matrix $\bW$. In all the experiments we simply threshold the square of the spectral distance and set the weights according to \ref{Thresh}:

\begin{equation}
	\label{Thresh}
	\left\{
	\begin{array}{ll}
		W_{ij} = 1  & \text{if } \|\bs_i - \bs_j\|_2^2 < d_{\min}^2  \\
		W_{ij}  = 0  & \text{otherwise},
	\end{array}\right.
\end{equation}
where $d_{\min}^2$ represents the maximum squared spectral distance required in order to consider that two pixels are similar. As previously explained in Section \ref{sec:GraphMap}, there are different heuristics for choosing the weights. \eqref{Thresh} was sufficient in our experiments to demonstrate the effectiveness of the method. The most appropriate definition of $\bW$ is out of the scope of this paper. Figure \ref{TrueImage} shows the affinity matrix of Data$2$ when we have $\text{SNR}=20$ dB, $d_{\min}^2=1.8$. For the purpose of display the points have been re-ordered  in such a way that the pixels belonging to the $5$ squares of the first row appear first, then those of the squares of the second one appear second, and so on. The pixels of the background are moved to the end. Note that the affinity matrix has $6$ white blocks (corresponding entries $W_{ij}$ are $1$), the first white blocks indicates that all the squares on the row are similar. The algorithm described in \cite{Belkin2003} is then used to cut the graph into $10$ disjoint subgraphs and then unmixing is performed on each subgraph.

We compared the performances of FCLS \cite{Heinz01} and SUnSAL-TV \cite{Iordache2012} with the proposed approach denoted by GLUP-Lap (Group Lasso with Unit sum, Positivity constraints and graph Laplacian regularization). We used the Root Mean Square Error (RMSE) defined as $\text{RMSE} = \sqrt{\frac{1}{N  L} \times \| \boldsymbol{\hat{A}} - \bA\|_F^2}$ as the evaluation metric. We tested SUnSAL-TV and GLUP-Lap for different combinations of the sparsity and the spatial tuning parameters $\mu$ and $\lambda$. Table  \ref{table:one} reports the best performance of each algorithm for a given data set and a given SNR with the corresponding optimal pair of regularization parameters. GLUP-Lap requires the tuning of an additional parameter $d_{min}^2$ which is also reported in the table. Both, SUnSAL-TV and GLUP-Lap, outperformed FCLS. GLUP-Lap had the lowest RMSE for all cases. As the SNR increases the the rate at which GLUP-Lap improves with respect to FCLS increases. This is due to the fact that the observations contain less noise, thus the adjacency matrix becomes more reliable. The simulations performed with Data$2$ show that this data set is more difficult than the previous since it contains a large number of endmembers: $15$ compared to $5$ in Data$1$. As before, GLUP-Lap outperformed FCLS and SnSAL-TV. It is important to note that GLUP-Lap and SUnSAL-TV were run under the same ADMM conditions. The penalty parameter was set to $0.05$, and the maximum number of iterations to $200$ in both algorithms.

The first row of Figure \ref{AbdMaps} shows the true abundance map of endmember $\be_2$ in Data$1$, and the estimated maps obtained with FCLS, SUnSAL-TV and GLUP-Lap with SNR=$30$dB. It can be seen from these maps that both SUnSAL-TV and GLUP-Lap estimated smooth abundance maps compared to FCLS. Note that the squares that were not correctly estimated by SUnSAL-TV were better estimated with GLUP-Lap. This is possibly due to the fact that these squares are similar, for this reason they were encouraged to have similar estimates and appeared as consistent blocks in the abundance map estimated by GLUP-TV. The same observation can be made in the second row of Figure \ref{AbdMaps} that shows the abundance map for $\be_{11}$ in Data$2$ with SNR=$30$dB. FCLS was not able to correctly estimate the abundance of this endmember in the figure. SUnSAL-TV, possibly due to the links it form with its surrounding estimates also failed to correctly estimate the abundances of $\be_{11}$. despite the difficulty of this abundance map, GLUP-Lap perfectly recovered the abundances. Even if the $5$ squares are separated by the background, the corresponding pixels are connected in the graph due to their similarity and collaboratively estimate their abundances.  

\vspace{-0.5cm}
\begin{table}[h]
\caption{RMSE obtained with different values of the SNR, with the optimal values of the couple ($\mu$; $\lambda$) for SUnSAL-TV and GLUP-Lap, the penalty parameter was set to $\rho = 0.05$ for both algorithms.}
\begin{center}
\begin{tabular}{ c | c | c | c }
\rowcolor{Gray}	& SNR $20$ dB & SNR $30$ dB & SNR $40$ dB \\ \hline 
\rowcolor{LightCyan} \multicolumn{4}{c}{Data1}\\ \hline 
FCLS		& $0.262$ & $0.0173$ &	$0.0101$ \\ \hline 
SUnSAL 	         & $0.0165$ & $0.0101$ & $0.0031$	\\ 
TV		         & ($5~10^{-4}$; $0.05$) &($5~10^{-4}$; $0.01$) & ($5~10^{-4}$; $0.05$)			\\ \hline
GLUP                & {$0.0152$} &	$0.0049$ & $0.0012$\\ 
Lap	                 & ($0.01$; $0.5$) & ($5~10^{-4}$; $0.5$) &	($5~10^{-5}$; $0.5$)		\\ 
			& $d_{min}^2 = 2.5$ & $d_{min}^2 = 0.3$ & $d_{min}^2 = 0.05$ \\ 
\hline
\rowcolor{LightCyan} \multicolumn{4}{c}{Data2}\\ \hline 
FCLS		& $0.0307$ & $0.0240$ &	$0.0151$ \\ \hline 
SUnSAL 	         & $0.0250$ & $0.0132$ & $0.0073$	\\ 
TV		         & ($0.05$; $0.3$) &($10^{-4}$; $0.005$) & ($5~10^{-5}$; $10^{-3}$)			\\ \hline
GLUP                & $0.0174$ &	$0.0078$ & $0.0023$\\ 
Lap	                 & ($0.01$; $1$) & ($10^{-4}$; $1$) &	($5~10^{-5}$; $1$)		\\ 
			& $d_{min}^2 = 1.8$ & $d_{min}^2 = 0.5$ & $d_{min}^2 = 0.5$ \\ \hline

\end{tabular}
\label{table:one}
\end{center}
\end{table}

\vspace{-1cm}
\section{Conclusion}
\label{sec:conclusion}
In this work we used the affinity matrix of the image in order to incorporate the graph Laplacian regularization within the sparse unmixing formulation. We showed that the resulting graph regularized framework has potential in improving the abundances' estimation accuracy and creates more consistent areas at the local and global level. Future work includes validating the proposed approach using real data sets and studying the potential of spatial-spectral weights for further improving the performance of the proposed approach.

\bibliographystyle{IEEEbib}

\bibliography{ref}

\end{document}